\documentclass[letterpaper, 10pt, conference]{ieeeconf} 
\IEEEoverridecommandlockouts
\usepackage{graphicx}
\usepackage{subcaption}
\usepackage{amsmath,amssymb}
\usepackage{algorithmic}
\usepackage{algorithm}
\usepackage{xcolor}
\usepackage{tabularx}
\usepackage{booktabs}
\usepackage{multirow}
\usepackage{orcidlink}

\usepackage[nolist,nohyperlinks]{acronym}
\usepackage{hyperref}
\hypersetup{
    colorlinks=true,
    linkcolor=blue,
    citecolor=teal,
    filecolor=magenta,      
    urlcolor=cyan,
    pdftitle={main},
    pdfpagemode=FullScreen,
    }
\usepackage[
backend=biber,
style=ieee,
sorting=none,
abbreviate=true,
doi=false,
maxbibnames=3
]{biblatex}

\begin{acronym}
    \acro{VAE}{Variational AutoEncoder}
    \acro{CVAE}{Conditional VAE}
    \acro{MSE}{Mean Square Error}
    \acro{MAE}{Mean Absolute Error}
    \acro{KLdiv}{Kullback–Leibler divergence}
    \acro{SMAPE}{Symmetric Mean Absolute Percentage Error}
    \acro{WMAPE}{Weighted Mean Absolute Percentage Error}
    \acro{DoFs}{Degrees of Freedom}
    \acro{CNN}{Convolutional Neural Network}
    \acro{MLP}{Multi Layer Perceptron}
    \acro{PCA}{Principal Component Analysis}
    \acro{t-SNE}{t-Distributed Stochastic Neighbor Embedding}
    \acro{RMSE}{Root Mean Square Error}
\end{acronym}

\title{\LARGE \bf Towards Interpretable Visuo-Tactile Predictive Models\\for Soft Robot Interactions}
\author{Enrico Donato$^{1}$ \orcidlink{0000-0002-8844-5279}, Thomas George Thuruthel$^{2}$ \orcidlink{0000-0003-0571-1672} and  Egidio Falotico$^{1}$ \orcidlink{0000-0001-8060-8080}
    \thanks{This work received funding from the European Union’s Horizon 2020 research and innovation program under grant agreement No. 863212 (PROBOSCIS project) and the Royal Society research grant RGS/R1/231472.}
    \thanks{$^{1}$E. Donato and E. Falotico are with The BioRobotics Institute, Sant'Anna School of Advanced Studies, 56025 Pontedera (PI), Italy and with the Departement of Excellence in Robotics \& AI, Sant'Anna School of Advanced Studies, 56125 Pisa, Italy {\tt\small \{e.donato, e.falotico\}@santannapisa.it}}%
    \thanks{$^{2}$T.G. Thuruthel is with the Department of Computer Science, University College London, London, United Kingdom {\tt\small t.thuruthel@ucl.ac.uk}}%
}

\begin{document}
    \maketitle   

    \begin{abstract}
        Autonomous systems face the intricate challenge of navigating unpredictable environments and interacting with external objects. The successful integration of robotic agents into real-world situations hinges on their perception capabilities, which involve amalgamating world models and predictive skills. Effective perception models build upon the fusion of various sensory modalities to probe the surroundings. Deep learning applied to raw sensory modalities offers a viable option. However, learning-based perceptive representations become difficult to interpret. This challenge is particularly pronounced in soft robots, where the compliance of structures and materials makes prediction even harder. Our work addresses this complexity by harnessing a generative model to construct a multi-modal perception model for soft robots and to leverage proprioceptive and visual information to anticipate and interpret contact interactions with external objects. A suite of tools to interpret the perception model is furnished, shedding light on the fusion and prediction processes across multiple sensory inputs after the learning phase. We will delve into the outlooks of the perception model and its implications for control purposes. 
    \end{abstract}

    
    \section{Introduction} \label{sec:introduction}
        Deploying autonomous systems in the real world requires a thoughtful consideration of robots as embodied agents. Their physical form plays a crucial role in shaping dynamic interactions with the external environment \cite{gil2016perception}. Most importantly, this includes how the agent perceives the world and constructs a model upon which goal-based decision-making will be predicated. In particular, soft robotics arises as a challenging domain when building perception models, because of robot structural and material compliance making sensory prediction hard to refine \cite{pique2022cl,bianchi2023softtoss}.
        
        Biological perception frequently relies on the integration of multiple modalities, enabling the acquisition of a comprehensive set of features related to the same perceived event. This integration serves to disambiguate information and facilitate cross-inference among different sensory modalities \cite{proulx2012perceplearning}. Such biological perspective can be translated to robotic agents, whose sensing capabilities should both provide feedback about the control actions outcome and eventual interactions with the external environment. It does not produce a model of itself as an isolated body that has to deal with interactions as disturbances, but takes advantage of interactions to produce a world model in which the robot is aware of its surroundings \cite{soter2018bodilyaware}. Multi-modality is not the only challenge while dealing with perception, but sensory distribution plays a fundamental role, especially in the presence of continuously deformable bodies, whose relative parts deformation cannot be inferred by only localized sensing \cite{hedge2023sensingsoft}.
        
        Addressing the complexities of multi-modal distributed perception presents several challenges, encompassing issues like encoding, sensory coherence, confidence across modalities, and the intricate process of fusion \cite{lahat2015fusion}. The high-dimension data poses a formidable obstacle, potentially restricting the efficacy of data-driven algorithms in discerning tangled patterns and understanding input-output dependencies. A well-defined methodology might open avenues for more effective utilization of diverse sensory inputs and facilitate the creation of sophisticated models capable of comprehending their interplay. The use of such models in learning-based controllers of soft robots \cite{laschi2023learningcontrol} will unveil a plethora of novel applications. 

        \begin{figure}[tb]
          \centering
          \includegraphics[width=\linewidth]{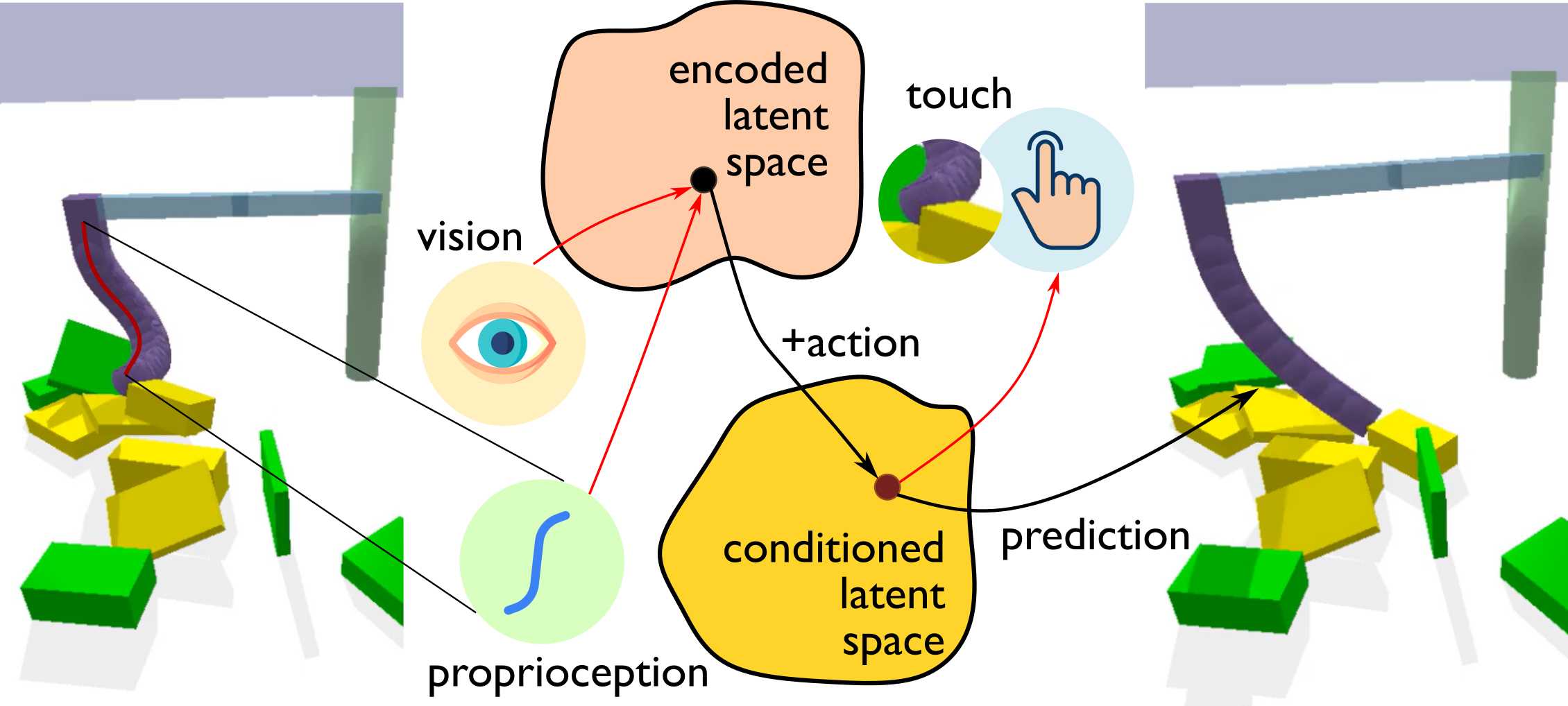}
          \caption{The soft finger interacts with the environment and multi-modal information is gathered and mapped in a shared latent representation. After being conditioned by the robot's future action, the latent representation is mapped back into the multi-modal sensory domain to get the predicted state.}
          \label{fig:fig1}
        \end{figure}

        In our previous work \cite{donato2024perceptiongenerative}, we have explored generative models to build a perception representation for soft robot interactions, as shown in Fig. \ref{fig:fig1}. In particular, a \ac{VAE} \cite{kingma2013autoencoder} is used to manage the multi-sensory encoding and fusion, and at the same time leverage its predictive capabilities to build a perception model, that aims to reconstruct sensory information via cross-inference. Such mapping is enabled by the projection of the multi-modal sensory information into a latent space, an organized, rich, coherent representation of the world in a virtual domain, and provides a methodology to choose the next action based on the prediction of the future observation of the world. The deployment of the proposed perceptive model to a robotic agent would allow the prediction of contact forces just relying on visual and proprioceptive information, without the need for dedicated contact sensorization and avoiding related hardware drawbacks and computational loads.
        
        In this work, we aim to get a deeper understanding of the perception model, making the latent representation more interpretable. We have improved the architecture to have modality-specific encoders and decoders to best unveil specific sensory information, allowing for late fusion and early decoding capabilities. Our main contribution is to provide some tools and guidelines to: (i) explain how the latent space is organized to encoded modalities, and how such organization influences the next sensory state prediction; (ii) discuss how different modalities might influence the predictive performance and in which amount; (iii) exploit the generative properties of the model to span over a virtual system evolution, e.g. for action exploration.

        The construction of a generative world model is detailed in Sec. \ref{sec:worldmodel}, followed by a definition of a methodology to interpret the learnt sensory representation in Sec. \ref{sec:representation}. Sec. \ref{sec:results} provides a comprehensive presentation of the results and delves into a detailed discussion of the findings. A conclusive summary of the work, along with insights and future directions, is presented in Sec. \ref{sec:conclusion}.

    \section{Related works}
        \subsection{Multi-modal sensory fusion in robotics}
            Sensory modalities exhibit significant variations in dimensionality, data distribution, and sparsity \cite{duan2022multimodalsensors}. This diversity keeps holding within the field of robotics. In addition to meeting task-specific requirements, the sensorization of robots primarily focuses on capturing the position and movements of different body parts (i.e., proprioception). It also involves gathering information about the interaction with the environment, either locally or globally, through tactile and visual sensing \cite{babadian2023fusion}. While both vision and proprioception offer a continuous flow of information, albeit in different formats, touch is inherently sparse due to the intermittent nature of contacts. This wide range of characteristics highlights the challenge of multi-modal fusion, emphasizing the need to integrate each piece of information into a cohesive state estimate \cite{piechocki2023latentrepresentation}. In alternative, other methodologies aim to find a mapping across different modalities, focusing on cross-modal reconstruction capabilities \cite{martin2022recursiveestimation}.

            Multi-modal fusion is usually either estimated via Kalman \cite{chen2023anthrofusion} or Particle filters \cite{lu2017particle}, which recursively update an estimate of the state and find the innovations driving a stochastic process given a sequence of observations, but differ in the way they accomplish this goal, respectively by linear projections or by a sequential Monte Carlo method. Despite their good performance, they rely on analytical dynamic models that may be difficult to obtain in many cases, as well as the construction of effective proposal distributions \cite{chen2023differentiablefilters}. Learning-based solutions instead face this challenge and infer the models from data distributions. Hybrid solutions are differentiable filters, that provide a way to learn models end-to-end while retaining the algorithmic structure of recursive filters \cite{lee2020difffilters}. This can be especially helpful when working with sensor modalities that are high dimensional and have very different characteristics. However, traditional data fusion methods deal with multi-modal data containing abundant inter- and cross-modality information, as in the case of deep learning \cite{tang2023fusiondeeplearning}. In particular, generative algorithms learn representations and formulate inference spaces while considering the complexity and reducing the redundancy of heterogeneous data, but at the same time performing such learning without supervision. The essential label information is available through inference about implicit relationships, making it clear that generative models potentially provide advantages in multi-sensory fusion. 
            
        \subsection{Interpretable multi-modal representation learning}
            Diverse information from various sensory sources introduces the challenge of effectively summarizing multi-modal data by leveraging their complementary and redundant aspects \cite{baltrusaitis2018multimodal}. This also raises issues regarding handling disparate levels of noise and addressing missing data. The objective of multi-modal representation learning is to discover a unified representation that conforms to the requirements of being a robust representation, as outlined by \textit{Bengio et al.} \cite{bengio2013representation} and \textit{Srivastava et al. }\cite{srivastava2014dropout}. Within this context, the \ac{VAE} seeks to generate a representation of the multi-modal input in a latent space, utilizing it for generative purposes by projecting it back into the multi-modal domain.

            The efficacy of generative models in this domain necessitates an exploration of how they instil confidence, facilitate decision-making, establish a foundation for assessment, or intrinsic purposes \cite{ross2021evaluating}. Posthoc techniques, including visualizations, have been devised to interpret the outputs of generative models based on deep learning. These techniques have been widely applied across various input modalities such as images, natural language, and domain-specific languages, as well as tabular data \cite{sercu2019peptide, vig2024bert}. Furthermore, efforts have been made towards developing disentangled representations that establish mappings between high-dimensional inputs and low-dimensional representations. The objective is to align representation dimensions with the underlying factors that generated the data \cite{chen2018disentanglement}. Additionally, such explanation often involves examining the variation in each dimension of the representation \cite{chen2016infogan}. While alternative solutions exist, they often lack insights into the individual meanings of dimensions and are typically specific to the encoding stage rather than the generative phase \cite{maaten2008tsne}. Although interactive methods have been devised to assist users in comprehending the data's geometry and understanding what information the representation retains, these methods fall short of explaining the dimensions of variation themselves.
                       
    \section{World Model Generation} \label{sec:worldmodel}
        In this section, we elucidate the process of data collection while a soft passive finger interacts with movable objects in simulation and multi-modal information is stored. Data are then used to train the predictive model, to build a world model that the robot might use for further decision-making or control purposes.
        
        \subsection{Simulation scenario}
\begin{figure}[tb]
              \centering
              \includegraphics[width=\linewidth]{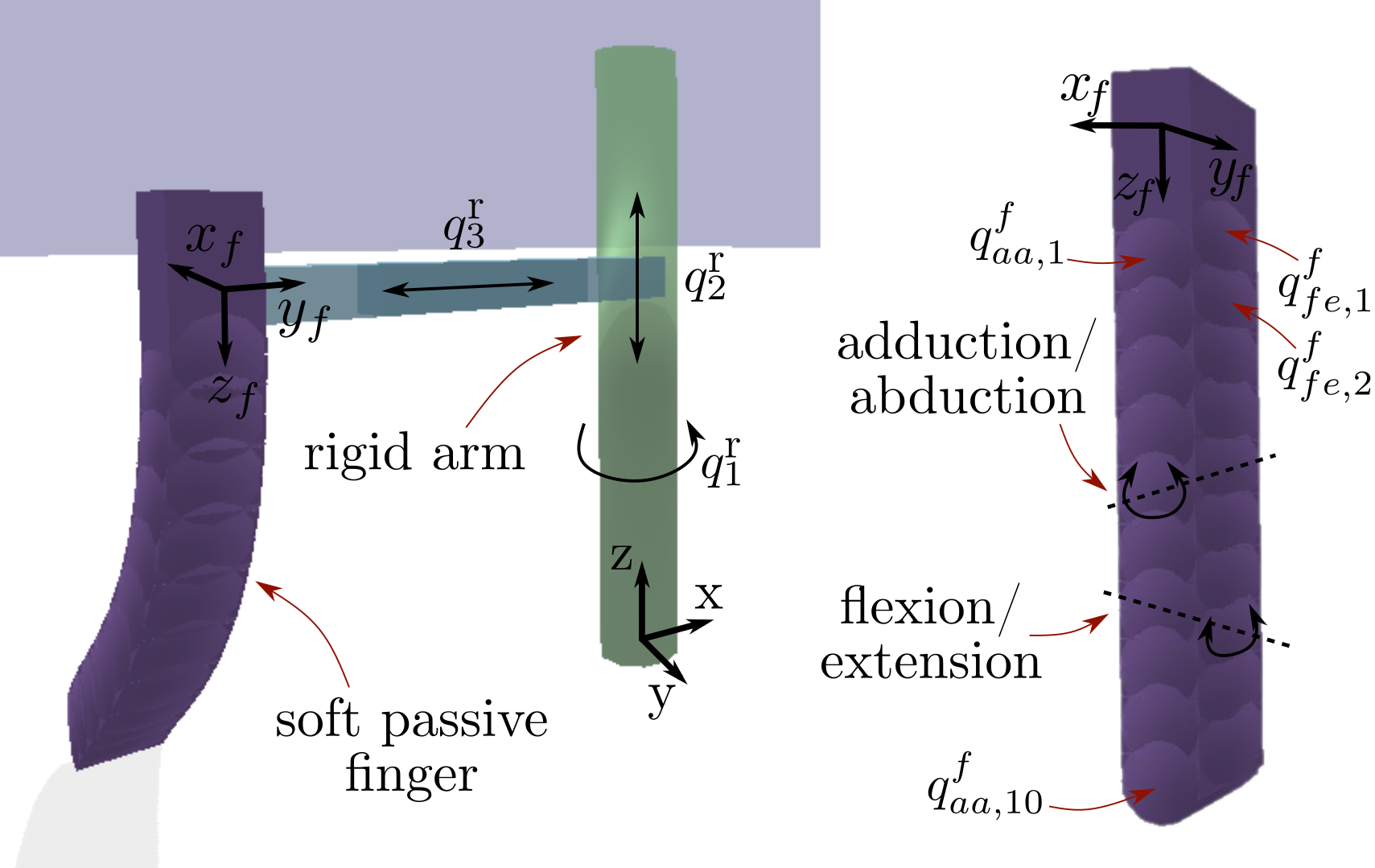}
              \caption{The simulation of the robotic platform from \cite{donato2024perceptiongenerative} involves incorporating a passive finger affixed to the distal portion of a rigid cylindrical robot. This finger interacts with the ground or potentially movable objects, exhibiting 20 \ac{DoFs} and executing flexion/extension as well as adduction/abduction movements.}
              \label{fig:fig3}
            \end{figure}

            The finger in Fig. \ref{fig:fig3} is implemented in the SoMo simulator \cite{graule2021somo} and comprises 20 links and joints, with the same spring constant and mass. Joints are arranged alternately from the base to the tip, allowing the finger to flex forward/backwards and laterally. This soft passive finger is attached to the distal end of a cylindrical rigid robot, featuring a rotary joint $q_1$ at the base and prismatic joints $q_2$ and $q_3$ connecting its links. This design provides a cylindrical workspace achieved through rotation and vertical and sliding motions. Box-shaped objects are introduced into the simulation, randomly placed near the robot, allowing the finger to make extra contact with its surroundings.
            
            The simulations start with a predetermined robot configuration, and random actuations are generated over defined time steps to cover the entire workspace. The simulation operates at 1 KHz to mitigate numerical instabilities. Data modalities are stored, including proprioceptive, contact, and visual information. The data acquisition frequency is set to 10 Hz, ensuring quasi-static motion. Proprioceptive data includes finger joint angles $q_t^f = {q_{fe,t}^f, q_{aa,t}^f} \in \Re^{20 \times 1}$ and rigid arm joint angles $q_t^r \in \Re^{3 \times 1}$, to capture the finger's shape and arm configuration.
            
            During interactions, SoMo monitors local deformations with normal forces $f_t \in \Re^{20 \times 1}$ on each finger link. Global deformations are captured visually through $v_t$ with shape $64 \times 64 \times 3$, obtained by recording the simulation with a virtual camera, offering a broader perspective on the system's deformations and interactions.

        \subsection{Perception model}
            Our learning architecture aims to seamlessly integrate information from diverse sensory modalities, leveraging the dynamic evolution of the physical body. In \cite{donato2024perceptiongenerative}, we employed a \ac{CVAE} \cite{sohn2015cvae} to predict the next state $\hat{s}_{t+1}$ in self-supervision at each time-step $t$ from non-minimal, multi-modal sensory data $\bar{s}_t$ and robot action $a_t$. 

\begin{figure}[tb]
              \centering
              \includegraphics[width=\linewidth]{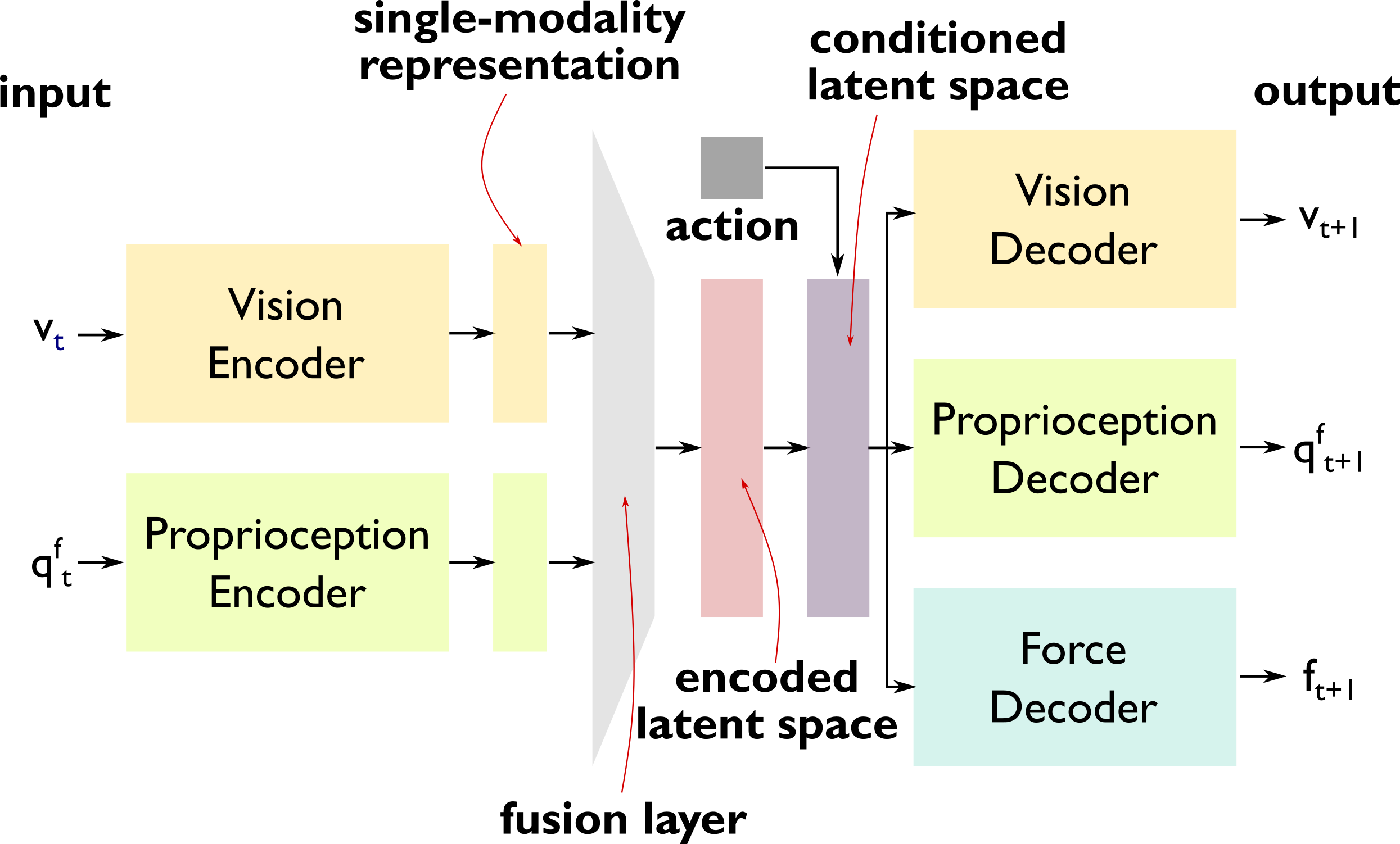}
              \caption{The perception model is implemented through a Conditional Variational AutoEncoder, with late fusion and early decoding stages thanks to modality-specific encoders and decoders. Fusion aims to map single-modality representations into the encoded latent space. Action conditioning enables the mapping of the information to the conditioned latent space, later used for sensory prediction.}
              \label{fig:fig2}
            \end{figure}

            In this study, we introduce an updated version of the generative model, accounting for the diversity of sensory data through modality-specific encoding and decoding layers, as illustrated in Fig. \ref{fig:fig2}. Each sensory input is processed by its encoder, generating a modality-specific model and mapping the information into a dedicated one-modality latent space. In particular, a \ac{CNN} is used for encoding/decoding of visual information, to consider also spatial features; conversely, two \ac{MLP} are employed for both proprioception and force mapping.
            
            Instead of fusing information at data level, fusion occurs at the decision level by mapping multiple single-modality latent spaces into a shared representation using a learned non-linear mapping, named \textit{encoded latent space}. This shared representation is then conditioned through the robot action and mapped to the \textit{conditioned latent space}, subsequently projected back into the multi-modal sensory domain using specialized decoders for each modality.

            During the training process, we optimize parameters for both the encoder and decoder networks using an ELBO loss. This loss considers both the reconstruction loss, minimizing the difference between desired output and generated data, and a regularization term, encouraging a normal latent space distribution.

    \section{Interpret the Sensory Representation} \label{sec:representation}
        This section explores a suite of tools designed for interpreting generative models. The methodology employed addresses the problem through two distinct approaches. First, it assesses the encoding and fusion stages by visualizing the organizational structure of the encoded and conditioned latent spaces concerning input modalities. Second, it scrutinizes the generative capabilities of the network by manipulating either the encoded representation or the selected action arbitrarily. 

        \subsection{Latent space visualization}
            Latent space denotes the learned, lower-dimensional space where a model encodes high-dimensional input data into a more concise and abstract representation. However, the challenge lies in gaining insights into the features of this high-dimensional latent space. One approach involves analyzing features individually or identifying evident patterns among sampled data. In the latter case, latent space visualization proves valuable in understanding how the model captures and organizes information, particularly aiding in data clustering and anomaly detection. This visualization maps the representation into a very low-dimensional embedding, typically two dimensions, making it easily interpretable.

            \ac{PCA} \cite{pham2022pca} identifies principal components that capture maximum variance in the data for linear dimensionality reduction, offering valuable insights into the overall structure of the latent space. However, identifying linear dependencies in such high-variable spaces is uncommon. In this context, \ac{t-SNE} \cite{maaten2008tsne} prioritizes preserving pairwise similarities between data points, making it well-suited for revealing the local structure of the latent space. It is worth noting that \ac{t-SNE} excels in preserving local structures, but it may not always accurately represent global structures. Anyway, data clusters often signify groups of similar instances or patterns within the data, shedding light on the distribution of points and providing insights into how the model captures features and variations. 
            
            The \ac{t-SNE} algorithm includes a tuning parameter known as \textit{perplexity}, which serves as a balance between local and global features. This parameter influences the effective number of neighbours that each point considers during dimensionality reduction. Through an iterative process of increasing the perplexity, the optimal value is identified based on achieving the lowest \ac{KLdiv} \cite{kimura2021tsne}, where a lower value indicates a reduced divergence between the latent space and a normal distribution.

        \subsection{Generative model properties}
            Upon gaining insights into the mapping of input sensory modalities into a shared latent space, the next step involves evaluating the generative capabilities of the network. While the network excels in predicting the acquired dataset optimally, as evidenced by the reconstruction loss, its true generative potential emerges when presented with synthetic data. In such instances, it adeptly leverages its generative capacities to establish meaningful connections between novel information and pre-existing knowledge.

            Assessing the generative properties of the perception model entails examining how the conditioned latent space is reconstructed during the decoding stage. Treating the decoders as constant mapping blocks, we can manipulate either the encoded latent representation or the action. In the former scenario, introducing a new sensory observation allows us to analyze how the network generalizes across diverse observations while keeping the action constant. Conversely, in the latter case, maintaining the observation as constant and varying the action enables us to explore the entire actuation space and anticipate all potential future observations resulting from changes in action.
            
            This analytical approach also provides insights into the network's stability concerning data distribution. Operating within a variational latent space, where the encoded latent representation is sampled from the learned distribution, assessing how different samples from the same distribution yield consistent predictions serves as a measure of stability.


    \section{Results} \label{sec:results}
        \begin{figure}[b!]
          \centering
          \includegraphics[width=\linewidth]{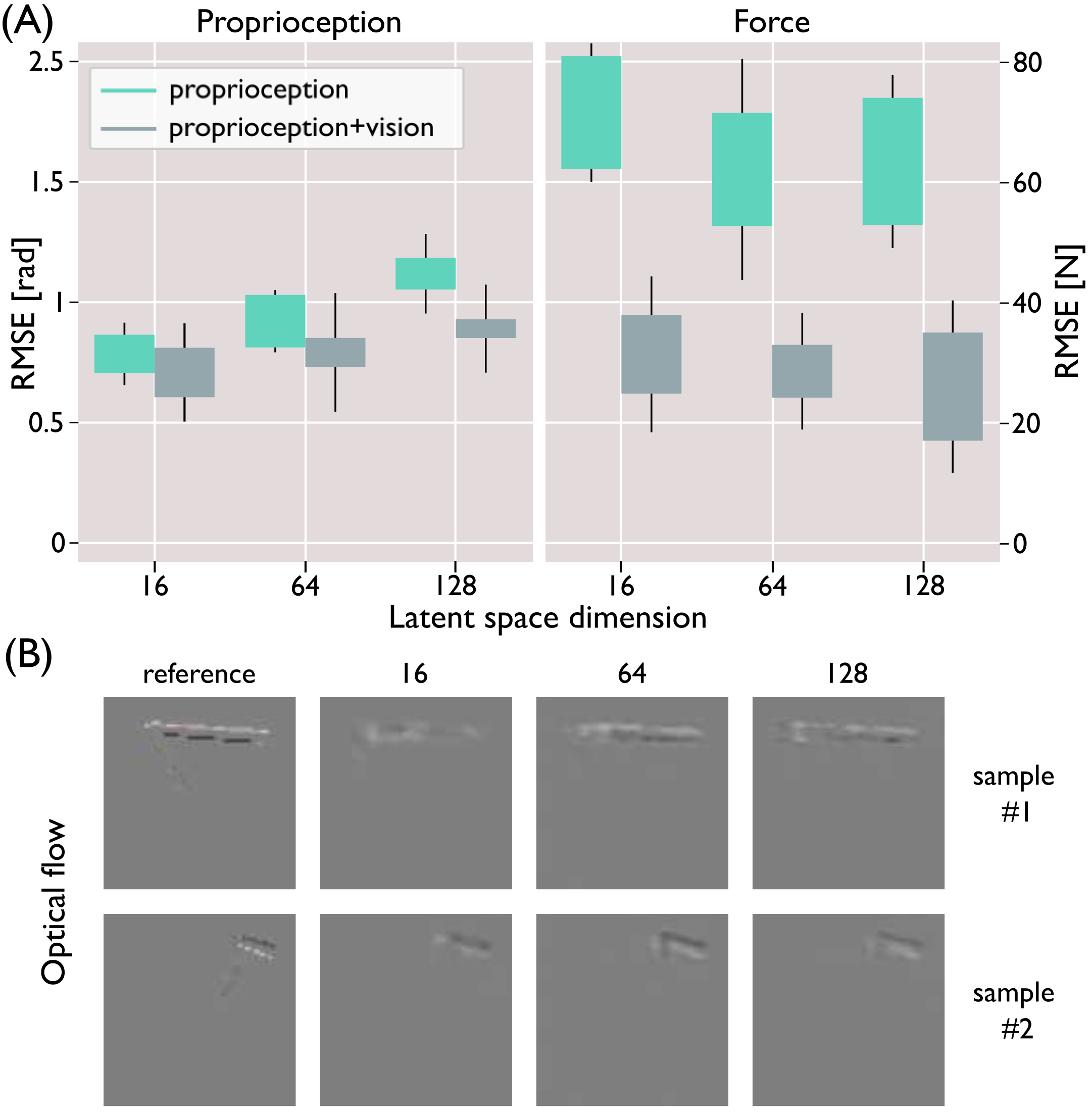}
          \caption{Prediction of the perception model over different output modalities. (A) Proprioception and force prediction over different latent dimensions and input modalities. (B) Optical flow prediction over different latent dimensions.}
          \label{fig:fig8}
        \end{figure}
        
        The examination of the generative architecture involves an assessment of its learning performance across various input modalities. It includes gaining insights into the organization of the latent representation in both the encoded and conditioned latent space. Furthermore, an analysis of generative capabilities will be conducted to evaluate prediction stability and observe the model's behaviour when presented with synthetic samples as input. Lastly, the consideration of input-to-output dependencies aims to provide an estimation of feature importance in modalities prediction.

        \subsection{Perception model performance assessment}
            The perception model is trained on a dataset comprising approximately 40k samples, using a laptop equipped with a Nvidia GeForce RTX 3060 GPU. The training process for the network takes between 20 minutes to 1 hour, depending on the size of the latent space. After training, each sensory prediction iteration averages 78 milliseconds.
            
            The assessment of the generative model performance extends to its predictive capabilities across consecutive time steps, as illustrated in Fig. \ref{fig:fig8}. Evaluation is conducted across various modalities, with a focus on force and proprioception depicted in Fig. \ref{fig:fig8}(A). The results align with previous discussions in \cite{donato2024perceptiongenerative}, quantified in terms of \ac{RMSE} between predicted values and the desired outcome. Notably, vision slightly enhances proprioception prediction, yet is reliable without any fusion, and enables a finer force estimation. Furthermore, expanding the latent space improves force prediction by accommodating a greater number of features from visual information, albeit with increased variance. 

            In Fig. \ref{fig:fig8}(B), optical flow predictions are showcased for two random samples from the validation dataset. Despite the suboptimal estimation of colours, the network adeptly identifies the image components responsible for detecting optical flow. However, improvements can be made in predicting the soft part, as its pronounced variability in motion poses challenges in accurately estimating changes in its shape. 
            
            Future deployment on physical robotic devices will then consider the presence of external cameras to track the body evolution and environment dynamics, as well as a proprioceptive system to estimate the robot shape over time. Contact sensors will serve as ground truth to measure the divergence between the model's predictions and real-world observations. However, these contact sensors can be removed during actual operational phases. The use of physical sensors introduces typical challenges associated with soft sensors, or to the quality issues encountered with visual information in sub-optimal settings. These sensory challenges will necessitate a pre-processing step before utilizing the perception model.
            
        \subsection{Latent space organization and visualization}
            Visualizing the latent space offers valuable insights into the underlying data distribution, forming the foundation for predictive modelling. This visualization is facilitated by the \ac{t-SNE} algorithm, with a perplexity parameter set to 1000. The choice of perplexity is informed by the observed monotonous decrease in divergence during its increase.

\begin{figure*}[tb]
              \centering
              \includegraphics[width=\textwidth]{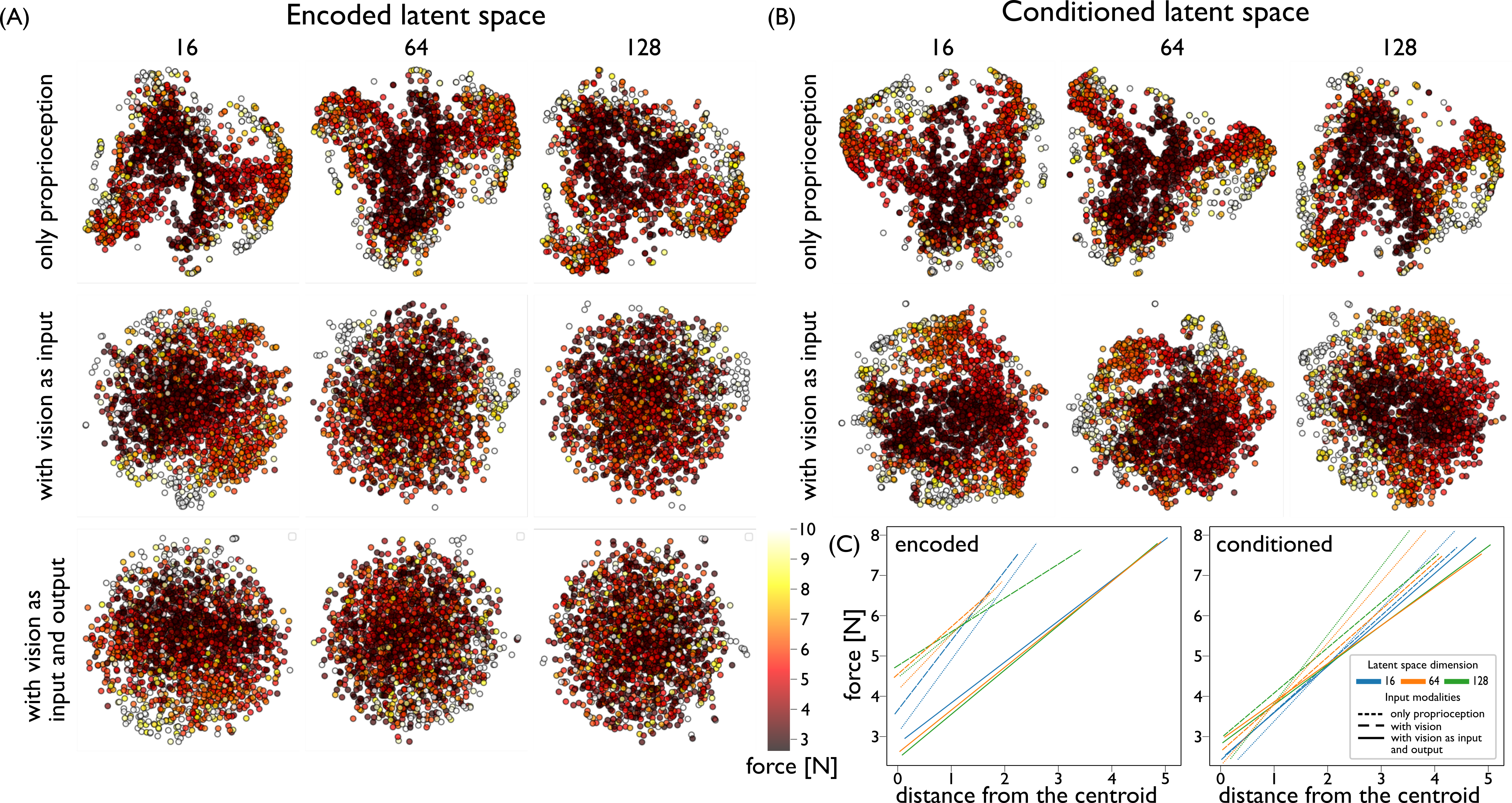}
              \caption{Latent space visualization and analysis. (A) Encoded latent space and (B) conditioned latent space over different modalities and latent dimensions. (C) Relation between distance from cluster centroid and force.}
              \label{fig:fig4}
            \end{figure*}

            In Fig.\ref{fig:fig4}(A), the encoded latent representation across two dimensions is presented. The analysis explores variable latent space dimensions and diverse input information. Each point in the bi-dimensional plane is color-coded based on the contact force applied to the finger at that specific timestep. When utilizing only proprioception, distinct clusters emerge to force magnitude, with symmetrical patterns apparent, especially around the axis of low forces (under 4 N). The centroid excels at encoding low-magnitude forces, while the outer region encompasses samples with higher applied forces. However, introducing vision as input disrupts this clarity, presenting a seemingly random distribution, particularly as the latent space dimensions increase.
            
            In contrast, Fig.\ref{fig:fig4}(B) demonstrates the impact of moving information into the conditioned latent space. The representation retains the symmetry property for proprioception-only input while reducing the spread of each cluster. For vision input, a discernible pattern of force distribution emerges, featuring a centroid with low forces that progressively increase towards the extremities. This relationship is further elucidated in Fig.\ref{fig:fig4}(C) by illustrating the distance between each sample and the distribution centroid concerning the applied force. The conditioned latent space allows for a spatially dependent encoding of force, a capability limited in the encoded latent space. These findings underscore the cross-modal inference capabilities of the perception model, facilitating the construction of a force-encoded representation even without explicit training for such information.

            An alternative metric to assess improvement upon transitioning to the conditioned latent space is the change in mutual information between the distance from the respective centroid and the applied force. Mutual information serves as a measure of dependence between two random variables, with higher values indicating increased dependence. The variation in mutual information as one traverses the latent space can be likened to an information gain. Examining Tab.\ref{tab1}, it becomes evident that the gain is nearly negligible when employing only proprioception, indicating that the transformation in the conditioned latent space maintains a similar level of topological information as its precursor. In contrast, when incorporating vision as input (or output), the information gain experiences a notable increase. This observation underscores how the inclusion of multi-modal information enhances the model's ability to discriminate among events, thereby amplifying its predictive capabilities.

            \begin{table}[]
                \begin{tabular}{ccccc}
                \multirow{2}{*}{\textbf{input modalities}}                                                  & \multicolumn{3}{c}{\textbf{latent space dimension}} & \multirow{2}{*}{\textbf{latent space}} \\ \cline{2-4}
                                                                                                            & 16              & 64              & 128             &                                        \\ \hline
                \multirow{2}{*}{only proprioception}                                                        & 0.32            & 0.33            & 0.30            & encoded                                \\ \cline{2-5} 
                                                                                                            & +9\%            & -3\%            & +7\%            & conditioned                            \\ \hline
                \multirow{2}{*}{with vision as input}                                                       & 0.17            & 0.07            & 0.04            & encoded                                \\ \cline{2-5} 
                                                                                                            & +88\%           & +285\%          & +650\%          & conditioned                            \\ \hline
                \multirow{2}{*}{\begin{tabular}[c]{@{}c@{}}with vision as \\ input and output\end{tabular}} & 0.11            & 0.04            & 0.04            & encoded                                \\ \cline{2-5} 
                                                                                                            & +91\%           & +350\%          & +250\%          & conditioned                            \\ \hline
                \end{tabular}
                \caption{Mutual information between the latent space topology and the force on the finger. Variations over latent dimensions and modalities are considered. The information gain among encoded and conditioned latent space is reported.}
                \label{tab1}
            \end{table}
            
        \subsection{Analysis of generative properties}
            Our analysis has primarily focused on the predictive capabilities of the perception model. Given its generative properties, it is pertinent to highlight its performance across various input modalities, actions, and when confronted with synthetic data. For brevity and clarity in subsequent discussions, the perception model with 64 latent units exploits proprioceptive and visual information to predict contact.

            \begin{figure*}[tb]
              \centering
              \includegraphics[width=\textwidth]{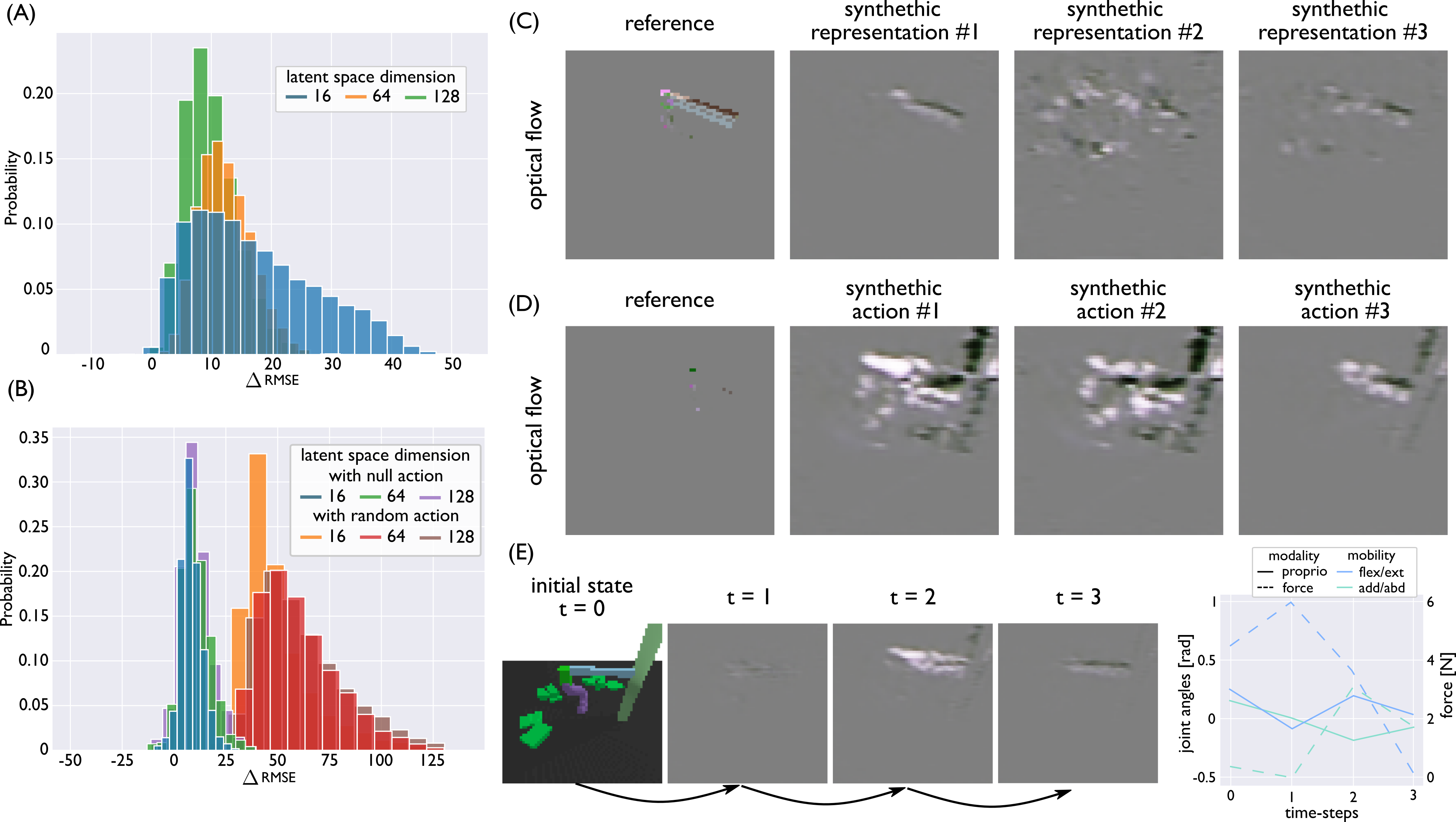}
              \caption{Generative model properties assessment. Variations in reconstruction capabilities result from (A) alterations in representation sampling from the encoded latent space and (B) utilization of null or random actions. Data generation through (C) synthetic observations or (D) synthetic actions. (E) Feedback-loop of predicted information to input modalities over time.}
              \label{fig:fig5}
            \end{figure*}

            Input modalities undergo encoding into a shared variational latent space, with stochastic parameters being learned for latent representation sampling. However, the stochastic nature of this process can influence the network's prediction capabilities. In Fig. \ref{fig:fig5}(A), the impact of repetitive sampling from the same encoded distribution on prediction error is depicted. The likelihood of obtaining a sub-optimal representation is higher in low-dimensional latent spaces.
            
            Similarly, the alteration of conditioning actions is explored for its effect on prediction. Fig. \ref{fig:fig5}(B) illustrates the change in prediction error while using null actions, to not induce any motion, and randomly sampled actions. Null actions have a relatively minor impact as they do not provide additional information to the network, resulting in outputs closely resembling the inputs. Conversely, random actions alter performance by leading to unexplored system evolutions.
            
            Visual representations in optical flow prediction further validate these findings for synthetic latent representations [Fig.\ref{fig:fig5}(C)] and actions [Fig. \ref{fig:fig5}(D)]. Despite incorporating white noise into the sample while keeping the other input constant, the network generates new data, offering novel insights in the robotic domain. For instance, assessing the stability of specific actions in future observations by examining changes in representation while holding the action, or exploring diverse system evolutions by maintaining the same representation while altering the action.
            
            When a subset of outputs is fed as input, the perception model enables feedback loops for self-prediction. Fig. \ref{fig:fig5}(E) illustrates this approach across three timesteps, incorporating visual information (combining optical flow with the initial visual state) and proprioception as feedback to predict force dynamics. This iterative feedback mechanism enhances the prediction of the network's future states.

    \section{Conclusion} \label{sec:conclusion}
        This work introduces a methodology for analyzing and interpreting perception models based on generative techniques. The application of multi-modal distributed perception in soft robotic systems has revealed intricate challenges related to sensory encoding, fusion, and state prediction. Expanding on our previous work with generative models, we investigate the mapping of sensory observations to latent representations and explore the dynamic interplay between sensory inputs and subsequent robot actions, delving into sensory prediction.

        We aim to extend the model application to include actuated soft robots. The overarching objective is to leverage this concise and versatile state representation to develop task-specific control policies that leverage world models.
        
    \printbibliography
\end{document}